%% file: main.tex
\documentclass[10pt,twocolumn,letterpaper]{article}

\usepackage{wacv}
\usepackage{times}
\usepackage{epsfig}
\usepackage{graphicx}
\usepackage{amsmath}
\usepackage{amssymb}

\usepackage{times}
\usepackage{soul}
\usepackage{epsfig}
\usepackage{graphicx}
\usepackage{amsmath}
\usepackage{amssymb}
\usepackage{balance}
\usepackage[table]{xcolor}
\usepackage{subcaption}
\usepackage{tablefootnote}
\usepackage{multirow}
\usepackage{pifont}
\usepackage{xcolor}
\usepackage{caption}
\newcommand{\cmark}{\ding{51}}%
\newcommand{\xmark}{\ding{55}}%

\newcommand\blfootnote[1]{%
  \begingroup
  \renewcommand\thefootnote{}\footnote{#1}%
  \addtocounter{footnote}{-1}%
  \endgroup
}

\makeatletter
\let\NAT@parse\undefined
\makeatother
\usepackage[numbers]{natbib}

\usepackage[pagebackref=true,breaklinks=true,letterpaper=true,colorlinks,bookmarks=false]{hyperref}

\wacvfinalcopy 

\def\wacvPaperID{****} 

\ifwacvfinal\fi
\begin{document}

\title{Leveraging Pretrained Image Classifiers for Language-Based Segmentation}

\author{David Golub\\
Stanford University\\
{\tt\small golubd@cs.stanford.edu}
\and
Ahmed El-Kishky\\
University of Illinois at Urbana-Champaign\\
{\tt\small elkishk2@illinois.edu}
\and
Roberto Mart\'in-Mart\'in\\
Stanford University\\
{\tt\small roberto.martinmartin@stanford.edu}
}

\maketitle

\begin{abstract}
Current\blfootnote{\url{https://sites.google.com/stanford.edu/cls-seg} This work has been supported by the ONR, MURI
award W911NF-15-1-0479.} semantic segmentation models cannot easily generalize to new object classes unseen during train time: they require additional annotated images and retraining.  We propose a novel segmentation model that injects visual priors into semantic segmentation architectures, allowing them to segment out new target labels without retraining. As visual priors, we use the activations of pretrained image classifiers, which provide noisy indications of the spatial location of both the target object and distractor objects in the scene.  We leverage language semantics to obtain these activations for a target label unseen by the classifier. Further experiments show that the visual priors obtained via language semantics for both relevant and distracting objects are key to our performance 
\end{abstract}
\vspace{-2em}

\input{1-intro.tex}
\input{2-rw.tex}
\input{3-method.tex}

\input{4-expsetup.tex}
\input{5-experims.tex}
\section{Conclusion}
\label{conclusion}

We presented a novel segmentation method for previously unseen labels. Through language and transfer learning, we can transfer visual priors from image classifiers to semantic segmentation models. The method to transfer information uses saliency maps from pretrained classifiers that indicate possible locations for the object to segment and as well as locations of possible distractors. We thoroughly study the impact of the semantics, saliency maps, and which class labels are most challenging for our proposed method. 



\balance
{\small
\bibliographystyle{ieee}
\bibliography{egbib}
}

\end{document}

%% file: 1-intro.tex
\section{Introduction}
Semantic segmentation is the task of annotating image pixels with labels that correspond to one or several classes of interest. Recently, end-to-end models based on fully convolutional neural networks (FCNs)~\cite{hong2015decoupled,long2015fully, chen2017rethinking,chen2018deeplab} have significantly improved accuracy for semantic segmentation, holding the state-of-the-art on multiple benchmarks~\cite{cordts2016cityscapes,everingham2010pascal,zhou2016semantic}. The caveat is, though, that training such models for a new class of interest requires large amounts of image annotations either in the form of pixels that correspond to the class in the image~\cite{long2015fully,chen2017rethinking} or as bounding boxes around the objects of the class~\cite{khoreva2017simple, dai2015boxsup}. These annotations have to be done manually and are thus costly to obtain. Due to the annotation burden, high-quality segmentation models only exist for a few sufficiently annotated classes. 

Previous methods have proposed two approaches to extend the success of FCN-based segmentation to other classes requiring less (or at least easier) annotations.

The first of these are \emph{weakly supervised} approaches; these class of methods try to reduce the complexity on the annotation process by leveraging image-level labels, a class label for an entire image~\cite{hong2015decoupled, Zhang2018DecoupledSN, redondo2018learning}.
The second family of approaches apply \emph{few-shot} semantic segmentation~\cite{shaban2017one,rakelly2018conditional,zhang2018sg}; these approaches alleviate the annotation burden by learning to segment out a target class with the support of a few additional images. However, both of these approaches fail to leverage the semantic similarity new unseen classes and previously seen classes and require the addition of new manually-labelled images.

\begin{figure}[t]
\includegraphics[width=0.98\linewidth]{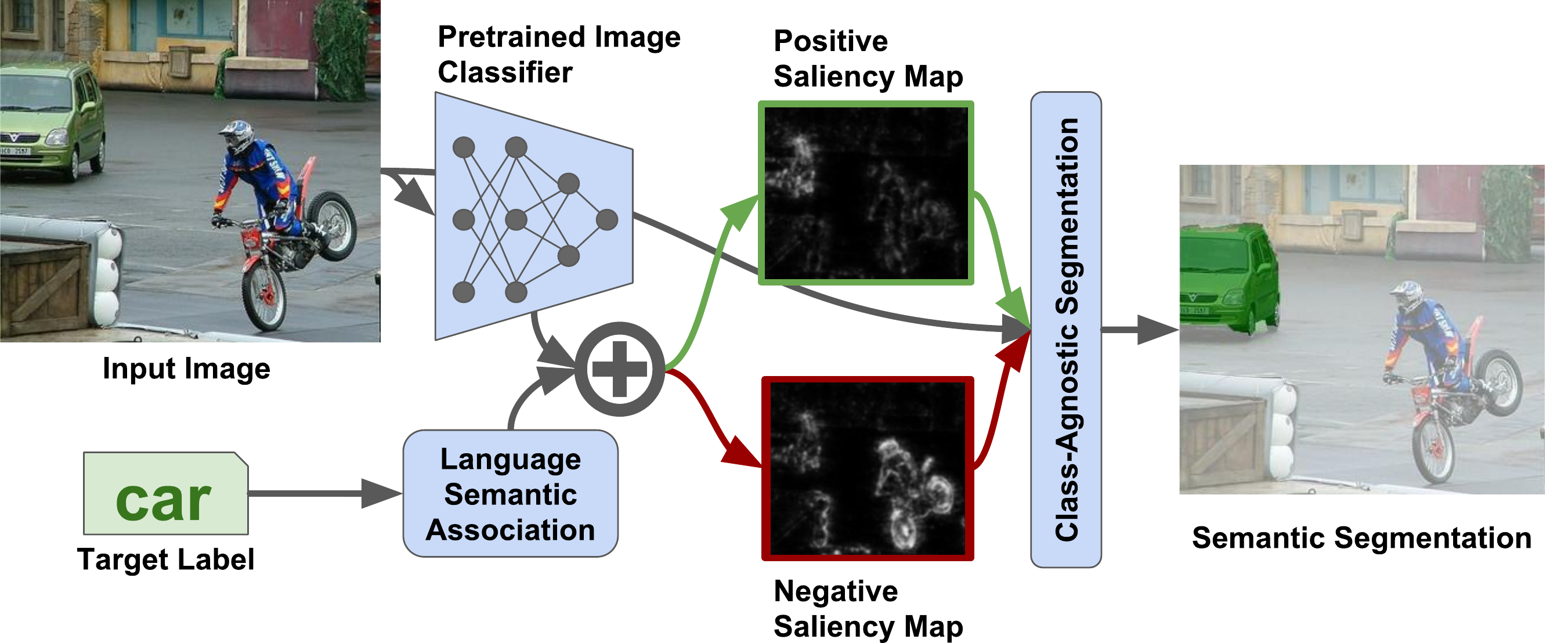}
\caption{Our approach for semantic segmentation; Given an image and a target label (\textit{car}), we leverage language semantics to extract visual priors from the pretrained classifier and generate a positive (green frame) and a negative (red frame) activation map, indicating regions of the image related and unrelated to the target; We use the activation maps as attention channels for a novel class-agnostic segmentation method that segments the original image}
\vspace{-5pt}
\label{fig:samples}
\end{figure}

\begin{table*}[t]
\footnotesize
\label{table:Comparison of methods}
\begin{tabular}{|l|l|c|c|@{}c@{}|c|c|}
\hline 
\multicolumn{2}{|l|}{Semantic Segmentation Method} & New Target Label & Annotation Level & Knowledge Transfer & Pretrained Classifier Role & Semantics Role \\
\hline
\hline
\multicolumn{2}{|l|}{Regular (e.g. \cite{long2015fully, lin2017refinenet,chen2018deeplab})} & {\color{red}\xmark} & Pixel & {\color{red}\xmark} & Initialize Weights & -\\
\hline
Weakly & \cite{ahn2018learning,Zhang2018DecoupledSN} & \multirow{2}{*}{{\color{red}\xmark}} & Image & \multirow{2}{*}{{\color{red}\xmark}} & \multirow{2}{*}{Initialize Weights} & Training Data \\
\cline{2-2}\cline{4-4}\cline{7-7}
Supervised & \cite{hong2015decoupled,siam2018rtseg, Kwak2017WeaklySS, dai2015boxsup} &  & Image and Pixel &  & & - \\ 
\hline 
\multicolumn{2}{|l|}{Few-Shot~\cite{rakelly2018conditional, shaban2017one,dong2018few,zhang2018sg}} &  {\color{green}\cmark}\rlap{~($>$1 an.)} & Image and Pixel & \begin{tabular}{c}from Classifier\\via Weights\end{tabular} & Initialize Weights & - \\
\hline
\multicolumn{2}{|l|}{Ours} & {\color{green}\cmark}\rlap{~(0 an.)} & Image and Pixel & \begin{tabular}{c} from Classifier via\\Language Semantics \end{tabular} &  \begin{tabular}{c}Initialize Weights and \\ Generate Saliency Maps \end{tabular} & Additional Input \\ 
\hline
\end{tabular}
\vspace{-5pt}
\caption{Comparison of families of semantic segmentation methods; Only our approach can segment semantically images for a new label without any additional annotation. All methods use pretrained classifiers to initialize weights, but ours use them also to generate saliency maps. Some weakly supervised segmentation models with image-level labels \cite{Zhang2018DecoupledSN,ahn2018learning} use saliency maps as training data while we use it as additional input} 
\label{table:weaklysupervised}
\vspace{-12pt}
\end{table*}

Recent work~\cite{zamir2018taskonomy} has shown that the representations learned by deep neural networks trained for different visual tasks are strongly correlated, and that transferring information between them is possible. 
Inspired by this idea, we aim to exploit a model trained for image classification for semantic segmentation of new or semantically-related classes.
Prior methods~\cite{Zhang2018DecoupledSN, redondo2018learning} have already shown that this transfer of information from classification to segmentation is possible. They use techniques such as Guided-Backprop~\cite{springenberg2014striving}, GradCAM~\cite{smilkov2017smoothgrad}, or SmoothGrad~\cite{selvaraju2017grad} to obtain noisy but interpretable object localizations in the form of saliency maps from image-level labels~\cite{ahn2018learning} (see Fig.~\ref{fig:samples}).
However, these methods train a new image classifier on the same dataset used for segmentation so that the labels for segmentation can also be back-propagated into the classifier. This procedure excludes them from the our setting since they cannot segment previously unseen, semantically related labels at test time, requiring retraining for each new label. 

In contrast, we propose to achieve semantic segmentation using off-the-shelf classifiers already trained on large image-level annotated datasets like ImageNet~\cite{deng2009imagenet}. These pretrained classifiers\footnote{We use the term \emph{pretrained classifier} as it is standard in the community for an off-the-shelf available model. However, we emphasize that we do not \emph{post-train} these classifiers with new data; our method uses them as-is.} are widely and easily accessible\footnote{A simple web search on ``pre-trained image classifier'' yields 19 variants of models pretrained on ImageNet.}. We use the activations of the neurons of a classifier to create saliency maps, noisy spatial indications of relevant areas in the image.

In our regime, our test-time target labels are not directly visible during training, although semantically similar concepts may have been seen by the off-the-shelf pretrained classifier. Therefore, we propose the use of language semantics to extract visual priors from the pretrained classifier, connecting the new unseen test-time labels to labels known by the pretrained classifier. We use an automatic WordNet-based~\cite{miller1995wordnet} process that maps the object class of interest (\emph{target label}) to a set of semantically-related labels contained in the classifier's training set that we can use to generate a saliency map. Saliency maps generated only from semantically related labels often include regions with distracting objects. Therefore, we also map the target label to adversary semantically unrelated labels present in the image (\textit{negative labels}) that generate additional \textit{negative} saliency maps. These two saliency maps contain spatial information of semantically related (positive) and unrelated (negative) objects in the image. Therefore, we attend on these saliency maps in a novel class-agnostic segmentation model (Fig. \ref{fig:samples}), that produces the final segmentation mask for the image.


Our full approach segments an input image for a given test label without any additional human annotation for labels inside or outside of the classifier's training set, and thus addresses the few-shot segmentation problem without retraining the model for unseen test-time labels. In summary, our main two contributions are as follows:
\begin{itemize}
\itemsep0em 
  \item We present a novel model for semantic segmentation that extracts visual priors from a pretrained image classifier using language semantics. Our model outperforms previous one-shot and five-shot methods of ~\cite{zhang2018sg} on the proposed partitions of Pascal VOC~\cite{shaban2017one} by 4\% and 3.2\% mIOU on Pascal VOC classes without additional human-annotated support sets at test time. 
  Our model also outperforms a foreground-background baseline and a model with a saliency map from a random label (no semantics) by 16.3\% mIOU. 
  \item We introduce a novel attention-based class-agnostic segmentation model that attends on two additional channels to the input image: a positive and a negative channel. The positive channel indicates regions possibly containing the target object while the negative channel indicates visually distracting areas, generated in our segmentation model from the activation of a pretrained classifier to related and unrelated semantic labels to the target label. Our model using positive and negative labels outperforms the best model using only semantically related labels by 8.4\% mIOU, with a gain of over 20\% mIOU on complex classes such as aeroplane, bus, car, dog, and bottle.
\end{itemize}

%% file: 2-rw.tex
\section{Related Work}
\label{rw}

Semantic segmentation is a long-standing task in computer vision, with applications in scene understanding~\cite{li2009towards,gupta2015indoor,socher2011parsing}, autonomous navigation~\cite{ess2009segmentation,nekrasov2018real,zhao2017icnet,siam2018rtseg}, and activity tracking~\cite{papoutsakis2013integrating,ondruska2016end}. Despite its longevity, this task remains challenging because object appearances and their visual boundaries vary widely depending on factors such as lighting, pose, scale and noise.

Recently, methods based on convolutional neural networks have shown impressive progress on this task~\cite{hong2015decoupled,Hu2017LearningTS, lin2017refinenet,long2015fully}. These networks are trained in a fully-supervised manner on large pixel-level annotated datasets such as Pascal VOC~\cite{everingham2010pascal}, City Scapes~\cite{cordts2016cityscapes} and MSCOCO~\cite{lin2014microsoft}. 
The dependency on thousands or more fine-grained, pixel-level annotated images is a limitation to generalize these models to new categories. 

Researchers have proposed two general procedures to alleviate the annotation burden to train semantic segmentation methods, \emph{weak-supervision} and \emph{few-shot} learning, that we review in the following.



\subsection{Weakly Supervised Semantic Segmentation} 
\label{rw:ws}

Weakly supervised (or partially supervised) semantic segmentation relaxes the pixel-level labeling dependency by using bounding boxes~\cite{khoreva2017simple, Hu2017LearningTS, ahn2018learning} or image-level annotations~\cite{Kwak2017WeaklySS, Zhang2018DecoupledSN, hong2015decoupled,chen2017deep} to generate automatic segmentation annotations for a target label. These annotations are then used to fine-tune a segmentation network that is originally trained on a densely annotated dataset.

Khoreva~\etal~\cite{khoreva2017simple} leverage bounding boxes to generate candidate segmentations for objects of interest. These segmentations are refined further with low-level vision techniques such as GrabCut~\cite{rother2004grabcut} and a class-agnostic semantic segmentation model. While the class-agnostic model and use of GrabCut is similar to our model, this technique still requires numerous bounding box annotations, in addition to a high-quality object detector trained on all target labels.

Several methods reduce the granularity of annotations further by training their own classifier with image-level class labels, and use them to generate saliency maps~\cite{Zhang2018DecoupledSN,redondo2018learning,hong2015decoupled}.  Two recent approaches ~\cite{Zhang2018DecoupledSN,redondo2018learning} use the saliency maps as noisy segmentation labels to finetune existing semantic segmentation models. Hong~\etal~\cite{hong2015decoupled} add a segmentation network on top of label-specific activations in the second-to-last layer of their trained classifier. This network is then finetuned on human-annotated segmentations of the target class. While conceptually related to our method, these approaches do not use pretrained classifiers but require a training process that assumes that all possible target labels are known during training. Moreover, once learned, these segmentation models need to be retrained for new labels and therefore cannot be applied to segmenting unseen labels at test-time. 

 

\subsection{Few-shot Semantic Segmentation}
\label{rw:fs}

Inspired by the recent success of few-shot and meta-learning approaches for object classification~\cite{long2018object, yang2018learning} and detection~\cite{chen2018lstd,rakelly2018conditional}, few-shot semantic segmentation methods reduce the annotation burden to several or sometimes a single annotation at pixel or bounding-box level for a new label. First introduced in~\cite{shaban2017one} and developed further in~\cite{rakelly2018conditional,zhang2018sg}, even for multiple labels at once in~\cite{dong2018few}, few-shot segmentation methods are usually based on conditional models that ground their predictions on a small set of pairs of images and segmentation masks. 

Zhang~\etal~\cite{rakelly2018conditional} propose a novel pooling strategy to leverage the similarity between the support set of segmentation masks and the target image. This method significantly outperforms prior work (5.2\% mIOU improvement). However, this method requires at least one human-annotated image at pixel-level with labels for each target, and cannot be applied using a test label only. 



%% file: 3-method.tex
\section{Semantic Segmentation Model}
\label{model}

\begin{figure*}[t]
\centering
\includegraphics[width=0.9\linewidth]{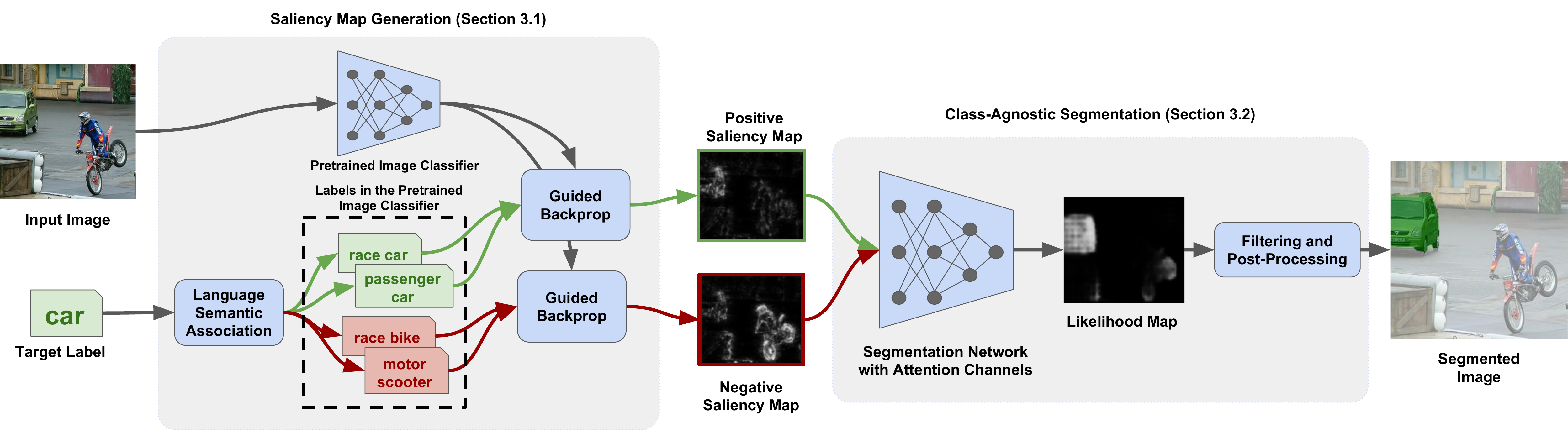}
\vspace{-7pt}
\caption{Diagram of our approach for semantic segmentation. At test time, our approach receives a new image and a previously unseen label (\textit{car}). We use a language semantic association to map the target to positive (green) and negative (red) labels in the set of training labels of the pretrained classifier. We then use Guided Backpropagation~\cite{springenberg2014striving} to extract saliency maps from the pretrained classifier corresponding to the positive and negative labels. The saliency maps are used as attention channels in a segmentation network to generate a likelihood image that is post-processed to generate the final segmentation mask of the unseen label}
\vspace{-15pt}
\label{fig:task}
\end{figure*}

The input to our method is a label of interest, $l$, and an RGB image, $I$, of width $w$ and height $h$. Our model generates a binary semantic segmentation mask, $M_{l} \in \{0,1\}^{ w \times h}$, indicating for each pixel if it belongs to an object of class label $l$ or not. 
We decompose the segmentation problem into two main steps as shown in Fig.~\ref{fig:task}. In the first step, we obtain a positive and a negative saliency map over the input image from an image classification network pretrained on ImageNet (Sec.~\ref{model:seg:heat}). The positive saliency map corresponding to a label is an image $S_l^{+} \in {R}^{w \times h}$ providing a noisy indication of the relevant image parts to that label. The negative saliency map is an image $S_l^{-} \in {R}^{w \times h}$ indicating image parts that are not relevant to the label, e.g. distracting objects.
Since in our segmentation task, the exact target labels are not part of the classifier's training set, we propose to exploit language semantics with WordNet to link the target label to related and unrelated labels seen by the pretrained classifier. 

In the second step, we feed the two saliency maps and the original input image into a 
semantic segmentation network with attention channels (a modified DeepLabv3~\cite{chen2017deep} architecture) that outputs a class likelihood image $P_{l} \in [0,1]^{w \times h}$. Each pixel $p \in P_{l}$ indicates the likelihood of the corresponding pixel in the original image to belong to the class label $l$ described by the saliency maps. Lastly, we post-process the likelihod image using GrabCut~\cite{rother2004grabcut} and obtain the final segmentation image $M_{l}$.

\subsection{Semantic Saliency Maps Generation}
\label{model:seg:heat}

\paragraph{Mapping target label to related and unrelated training labels:}

While pre-trained image classifiers can cover over several thousand object categories (also called \emph{synsets}), in our segmentation regime we assume that the exact target label was not seen during training, including the training of the classifier. We do, however, assume that labels semantically-related to the target label may have been seen by the pretrained classifier. As such, we propose to leverage language semantics to map the unseen target class to classes seen by the classifier during training.

Given a target label $l$ and an image classification model pretrained on a set of labels $L_{\text{train}}$, our goal is to map $l$ to $K$ semantically related labels $L^{+}_{sem}=[{l^{+}_{1},\ldots,l^{+}_{K}}]$, where $l^{+}_k \in L_{\text{train}}$. We use a language-semantics mapper to obtain these proxy labels. We will study two types of language-semantics mappers in Sec. \ref{model:grabcut}--WordNet~\cite{miller1995wordnet} and Word2Vec~\cite{mikolov2013distributed}. The language-semantics mapper can generate a very large number of proxy labels, larger than $K$. This would lead to a very noisy saliency map in the next step. Therefore, we further use the pretrained image classifier to prune the semantic proxy labels. We feed the image to the pretrained classifier and obtain a likelihood score for all proxy labels. We select among the resulting set of labels from the semantic mapper the $K$ with larger probability of be in the image as given by the image classifier.

Similarly, we would like to map $l$ to a set $L^{-}_{sem}$ of $K$ semantically unrelated labels that could act as distractors in the image. We assume that distractors are objects detected by the classifier that are not semantically related to $l$. Concretely, we use the $K$ labels of the classifier with highest likelihood score that are not part of $L^{+}_{sem}$.



\paragraph{Generating positive and negative saliency maps:}
Given our proxy positive label set ($L^{+}_{sem}$) or proxy negative label set ($L^{-}_{sem}$), we use class activation mappings~\cite{springenberg2014striving,smilkov2017smoothgrad,selvaraju2017grad} to extract the corresponding positive or negative semantic saliency map over the image, $S_l^{+}$ or $S_l^{-}$. Activation mappings indicate the areas in an image that most actively contribute to a certain classifier decision and are therefore correlated to the active class label. We will use the activation map as noisy location information about the object class.

Since the process is similar for positive and negative labels, we will discuss here for a generic label $l_k$ and a set $L_{sem}$. We first forward propagate the original image, $I$, through the pretrained image classifier. Then, we use each proxy label, $l_k$, as the target label and backpropagate the error signals back through the network to the input image to get a gradient tensor of size $\mathbb{R}^{3 \times w \times h}$. For each pixel location $i, j$ in the gradient tensor, we perform max pooling over the three channels and normalize the gradients over pixels so they lie in the range $[0, 1]$. This process produces the saliency map of a single positive or negative proxy label $l_k$: $S_{l_k} \in \mathbb{R}^{w \times h}$. 
Finally, we compute the average sum of the saliency maps from all proxy labels and generate the final saliency map $S_l \in \mathbb{R}^{w \times h}$. Repeating this procedure for $L^{+}_{sem}$ and $L^{-}_{sem}$ lead to the positive and negative saliency maps, $S_l^{+}$ and $S_l^{-}$. These saliency maps are passed as attention input to the next step, the class-agnostic semantic segmentation network.

\subsection{Class-Agnostic Semantic Segmentation with Attention}
\label{model:seg}

We propose to use the positive and negative saliency maps as spatial attention to guide a class-agnostic semantic segmentation model. To achieve that we modify a DeepLabv3~\cite{chen2017deep} architecture to condition the segmentation on the saliency maps, $S_l^{+}$ and $S_l^{-}$. The original DeepLabv3 architecture is a segmentation network $F: \mathbb{R}^{3 \times w \times h} \mapsto \mathbb{R}^{n \times w \times h}$, where $n$ is the number of classes. 
We concatenate the two saliency maps as additional channels to the original image and set $n=2$ (foreground or background) to obtain a modified DeepLabv3 network $F': {R}^{5 \times w \times h} \mapsto  {R}^{2 \times w \times h}$. The second channel of the output corresponds to the class likelihood image for the target label, $P_l$, while the first channel corresponds to the likelihood image of the background.

\begin{table*}[ht]
\centering
\begin{tabular}{|c | c | c |} 
\hline
{\small Target Label} & {\small WordNet Proxy Labels (ours)} & {\small Word2Vec Proxy Labels} \\
\hline\hline
{\small bottle} & {\footnotesize beer bottle, \textbf{bottle-screw}, pill bottle, soda bottle, water bottle } & {\footnotesize beer bottle, \textbf{bottle-screw}, pill bottle, wine bottle, soda bottle }\\ 
\hline 
{\small car} & {\footnotesize \textbf{car wheel}, racer, sports car, streetcar, \textbf{car mirror}, freight car } & {\footnotesize \textbf{car wheel}, racer, sports car, street car, pickup truck, fire truck }\\ 
\hline
{\small dog} & {\footnotesize pug, terrier, shepherd, tibetan terrier, \textbf{monkey dog}} & {\footnotesize terrier, pit bull, toy poodle, wire-haired fox terrier}\\ 
\hline
{\small chair} & {\footnotesize folding chair, barber chair } & {\footnotesize folding chair, barber chair, \textbf{desk}, \textbf{toilet seat}, \textbf{table lamp} }\\ 
\hline
{\small cat} & {\footnotesize tabby cat, tiger cat,  siamese cat, \textbf{cat bear}  } & {\footnotesize tabby cat, \textbf{fox terrier}, tiger cat, \textbf{toy terrier}, \textbf{hamster} }\\ 
\hline
{\small train} & {\footnotesize bullet train  } & {\footnotesize \textbf{school bus}, \textbf{freight car}, \textbf{streetcar}, steam locomotive }\\ 
\hline 
{\small sofa} & {\footnotesize studio couch } & {\footnotesize \textbf{folding chair}, \textbf{pillow}, \textbf{desk}, 
\textbf{bookcase}, studio couch }\\ 
\hline 
\end{tabular}
\caption{Top proxy ImageNet labels via our proposed WordNet-based method vs Word2Vec baseline; number of proxy labels is limited to $K=5$; words ordered by semantic similarity; \textbf{bold} indicates visually distracting proxy labels; Word2Vec links a significantly larger number of distracting labels}
\label{tab:one}
\vspace{-10pt}
\end{table*}

The final step is to obtain the binary segmentation mask, $M_l$, from the likelihood image, $P_l$. A direct way would be to threshold the likelihood. However, finding a single optimal binary threshold is hard because the mean and variance of pixel activations per likelihood image varies significantly across images. We propose instead to use GrabCut~\cite{rother2004grabcut} to produce our final segmentation mask. GrabCut uses four types of pixel-wise annotations: sure background $\textit{sb}$, probable background $\textit{pb}$, probable foreground $\textit{pf}$, and sure foreground $\textit{sf}$. We can map likelihood values to these four types of pixel annotations. We observe that the highest and smallest values in $P_l$, $p_{\text{\it max}}$ and $p_{\text{\it min}}$, vary per image and per label. Therefore, we design our annotations to be invariant to these shifts. Let $\delta = p_{\text{\it max}} - p_{min}$. Given threshold values $t_{\text{\it fg}}, t_{\text{\it bg}}, t_{\text{\it unk}}$, we create an annotation image $A$ where each pixel $A(u,v)$ is defined as:
\begin{align*}
\small
\begin{cases} 
      sg & \text{if  }  P_l(u,v) \leq p_{\text{\it min}} + t_{\text{\it bg}} * \delta \\
      pb & \text{if  } p_{\text{\it min}} + t_{\text{\it bg}} * \delta \leq P_l(u,v) < p_{\text{\it min}} + t_{\text{\it unk}} * \delta
       \\
      pf & \text{if  } p_{\text{\it min}} + t_{\text{\it unk}} * \delta \leq P_l(u,v) < p_{\text{\it min}} + t_{\text{\it fg}} * \delta
       \\ 
        sf & \text{if  } p_{\text{\it min}} + t_{\text{\it fg}} * \delta \leq P_l(u,v) 
   \end{cases}
\end{align*}
We use GrabCut with $A$ and obtain the final binary segmentation mask $M_l$.

%% file: 4-expsetup.tex
\label{model:grabcut}
\section{Experimental Setup}

\subsection{Implementation}

\paragraph{Pretrained Image Classifier:}
We use an image classifier pretrained on ImageNet1k as a pretrained model. Pretrained models on this dataset are the most studied and widely available. We use the VGG-19~\cite{simonyan2014very} architecture, as we observe its saliency maps are more concentrated on keypoints in the image as opposed to saliency maps from a ResNet50 architecture. 
We combine this classifier with Guided-Backpropagation~\cite{springenberg2014striving} to generate saliency maps.

\begin{table*}[t]
\centering
\resizebox{\textwidth}{!}{\begin{tabular}{|c|c|c|c|}
\hline
$i = 0$ & $i = 1$ & $i = 2$ & $i = 3$ \\
\hline\hline
\cellcolor{red!25}aeroplane, bicycle, bird, boat, bottle & \cellcolor{blue!25}bus, car, cat, chair, cow & \cellcolor{green!25}diningtable, dog, horse, motorbike, person & \cellcolor{yellow!25}potted plant, sheep, sofa, train, tv/monitor \\
\hline\end{tabular}}
\label{table:pascal_names}
\centering
\resizebox{1.0\textwidth}{!}{\begin{tabular}{l|cccc||c}\\
& \cellcolor{red!25} PASCAL-$5_{i=0}$ & \cellcolor{blue!25}PASCAL-$5_{i=1}$ & \cellcolor{green!25}PASCAL-$5_{i=2}$ & \cellcolor{yellow!25}PASCAL-$5_{i=3}$ & Mean\\ 
\hline\hline
OSLSM~\cite{shaban2017one}: 1-shot & 33.6 & 55.3 & 40.9 & 33.5 & 40.8 \\
coFCN~\cite{rakelly2018conditional}: 1-shot & 36.7 & 50.6 & 44.9 & 32.3 & 41.1 \\ 
SG~\cite{zhang2018sg}: 1-shot & 40.2 & 58.4 & 48.4 & 38.4 & 46.3 \\
\hline 
OSLSM~\cite{shaban2017one}: 5-shot & 35.9 & 58.1 & 42.7 & 39.1 & 43.9 \\
5-shot-coFCN~\cite{rakelly2018conditional}  & 37.5 & 50.0 & 44.1 & 33.9 & 41.4 \\ 
5-shot-SG~\cite{zhang2018sg} & 41.9 & 58.6 & 48.6 & 39.4 & 47.1 \\
\hline 
\hline
SEM-0-C-NONE: No saliency map & 39.6 & 40.3 & 37.4 & 31.6 & 37.2 \\
SEM-1-C-RAND: One saliency map from random label & 31.2 & 31.8 & 41.8 & 31.2 & 34.0 \\ 
SEM-1-C-GT: One saliency map from target label & 37.3 & 42.8 & 45.4 & 43.3 & 42.2 \\ 
\hline
SEM-2-C-RAND: Saliency maps from target label and random label & 40.8 & \textbf{57.9} & 47.7 & 38.5 & 46.2 \\ 
SEM-2-C-MEAN: Saliency maps from target label and top 5 labels & 43.1 & 56.2 & 47.12 & \textbf{47.0} & 48.4 \\ 
\textbf{SEM-2-C-NEG}: Saliency maps from target label and negative labels & \textbf{48.7} & 57.6 & \textbf{48.9} & 46.0 & \textbf{50.3} \\
\end{tabular}}
\caption{Mean intersection over union (mIOU) on the partitions described in ~\cite{shaban2017one}. We compare against the previously proposed one-shot and few shot models (OSLSM)~\cite{shaban2017one}, (FCN ~\cite{rakelly2018conditional}), (SG~\cite{zhang2018sg}). We then compare our proposed approach to a variant that only uses a single saliency map (SEM-1-channel-*), ablations without not using the generated saliency maps (SEM-0-C-NONE) and random labels (SEM-2-C-RAND, SEM-1-C-RAND) 
}
\label{table:full_results_k1}
\vspace{-10pt}
\end{table*}

\paragraph{Language Semantic Association:}

We implement and evaluate two possible semantic mappers:  WordNet~\cite{miller1995wordnet} and Word2Vec~\cite{mikolov2013distributed}. WordNet groups hierarchically words together by their semantic meaning. With WordNet we obtain first, a set of synonym labels, $[l_{1},\ldots,l_{N}]$, from the target label $l$, and then, a set of hypernyms for both the target label and the synonyms. Hypernyms are words that describe more general superclasses (categories) containing the target label and synonyms. The labels used to train image classifiers (e.g. ImageNet1K labels) are generic and often hypernyms of the labels used in image segmentation (e.g. Pascal VOC labels). We extract all labels from the classifier that have a word in common with any of the hypernyms and generate our set of semantic-related proxy labels.  

Word2Vec is a family of algorithms to generate word embeddings based on semantics. With Word2Vec we first embed the target labels $l$ and the labels in the set of possible proxy labels in a shared vector space using 300-dimensional GloVe embeddings~\cite{pennington2014glove} trained on the Common Crawl 840B word corpus.
For labels that contains multiple words, we average individual word embeddings into a single vector. Then, we choose the proxy labels with highest cosine similarity to $l$ as our set of semantic-related proxy labels. For both approaches, we set $K=5$ as the maximum number of proxy labels in our experiments. 


\paragraph{Modified DeepLabv3 with Attention Channels}

We use an open-source DeepLabv3 implementation 
and modify it as described in Sec.~\ref{model:seg}.  We use Resnet-50~\cite{he2016deep} to pre-initialize weights for all of the layers except the first and last layer. In the first layer, the weights for the first three channels are initialized using the pretrained ImageNet classifier. The weights for the fourth and fifth channel are randomly initialized. 

\paragraph{Conditional Fully Convolutional Network (baseline):}

We compare our work to a few-shot conditional Fully Convolutional Network (coFCN)~\cite{rakelly2018conditional} architecture that consists of five convolution and pooling layers, followed by one transpose convolution layer to generate the final predictions. The coFCN pools visual features from the annotated support set with outputs of a fully convolutional neural network. Then, these pooled features are run through a transpose convolution to produce final pixel-wise foreground/background probabilities. As in the original paper, we use VGG-16 to pre-initialize weights for all of the layers.

\paragraph{Post-Processing with GrabCut:}

We use the OpenCV implementation of GrabCut
and set $t_{fg}=0.7$, $t_{unk}=0.5$, and $t_{bg}=0.15$. We run GrabCut for five iterations.

\subsection{Dataset}

In our experiments, we use the target labels and images from the Pascal VOC dataset~\cite{everingham2010pascal}. The Pascal VOC dataset consists of roughly 11,000 images with 20 classes with a broad variety of object types (animals, vehicles, indoor/outdoor items, \ldots). 

As in prior work~\cite{shaban2017one}, we use the Semantic Boundaries Dataset~\cite{BharathICCV2011} and Pascal VOC train set~\cite{everingham2010pascal} to train our model, and use the Pascal VOC validation set for testing. We partition Pascal VOC into disjoint train/test label splits. Hence, in contrast to classical train/test partitions, our test examples share no labels in common with the train set.



We use the four label partitions from Shaban~\etal~\cite{shaban2017one} (Table \ref{table:full_results_k1}, top). In this setup, each partition consists of five to six consecutive labels, sorted in increasing alphabetic order. 



\subsection{Training}

The only part to train in our model is the class-agnostic semantic segmentation network with attention (Sec.~\ref{model:seg}). 
For this model, we use the default hyperparameters of the original DeepLabv3 for training. We use the model trained for 30,000 steps for testing.

%% file: 5-experims.tex
\section{Experiments}

\subsection{WordNet vs. Word2Vec Labels}
\label{ss:wnvsv}

We conduct five sets of experiments. In the first set, we compare the quality of proxy labels using a WordNet mapping to a Word2Vec embedding.  
We identified visually distracting labels as labels of objects that indicate a different class that can co-ocurr with the target label, and therefore, hinders the segmentation results. Table~\ref{tab:one} depicts some of the results of the comparison. The top three rows indicate cases when Word2Vec and WordNet extract similar proxy labels. The bottom four rows indicate cases when Word2Vec extracts visually distracting labels. WordNet may generate less than the maximum number of proxy labels, $K=5$, if the existing ImageNet1k labels are not related to the target label. 

A limitation of the Word2Net mapping is that it may miss semantically relevant words that are not in the WordNet ontology, e.g. the ImageNet label \textit{steam locomotive} from the label \textit{train}. Nevertheless, we find that Word2Vec consistently produces more visually distracting proxy labels than the WordNet-based approach. Only 64\% of the of the words extracted by Word2Vec are visually distracting as opposed to 18\% of the words extracted by WordNet. Based on this results, we use WordNet as the preferred semantic mapping in the rest of our experiments.

\subsection{Role of GrabCut} 
 In this experiment, we replace GrabCut postprocessing with a static threshold of 0.5 for the \textit{SEM-2-C-NEG} model. Without GrabCut, the performance drops to 48.1 mIOU, a 1.8\% improvement over the previous 1-shot state of art.

\subsection{Comparison Against Prior Work}
\label{ss:aolp}

In the second set of experiments, we evaluate and compare the accuracy of our proposed model to the performance of prior few-shot~\cite{shaban2017one,zhang2018sg} and one shot methods~\cite{rakelly2018conditional} on the proposed group partitions in~\cite{shaban2017one}. Note that in contrast to these methods, our method does not require any human-annotated images at test time; instead, we only use the label. We report mean intersection over union (mIOU) accuracy of our binary segmentation masks on the test set of Pascal-VOC.

\newcommand\figwww{0.80}
\begin{figure}[t]
\centering
\begin{subfigure}{\figwww\linewidth}%
\includegraphics[width=0.33\linewidth]{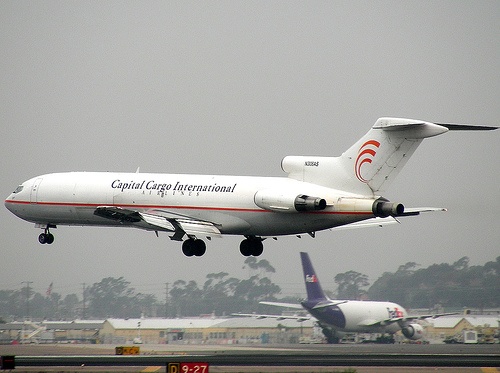}%
\hfill%
\includegraphics[width=0.33\linewidth]{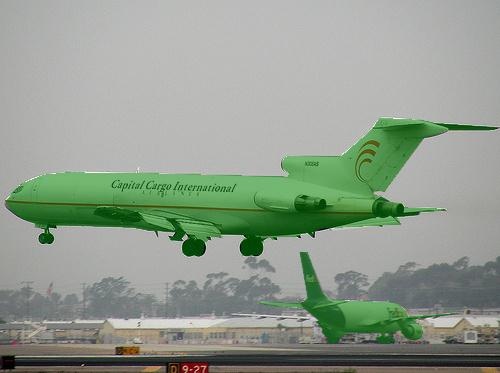}%
\hfill%
\includegraphics[width=0.33\linewidth]{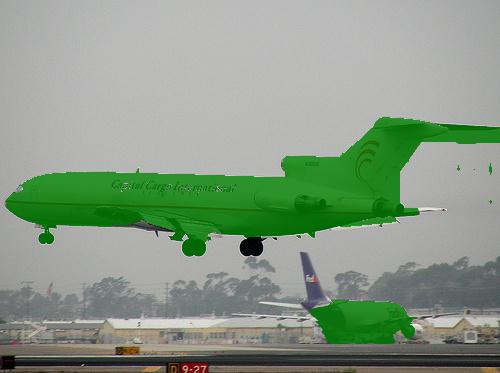}%
\end{subfigure}%
\\%
\begin{subfigure}{\figwww\linewidth}%
\includegraphics[width=0.33\linewidth]{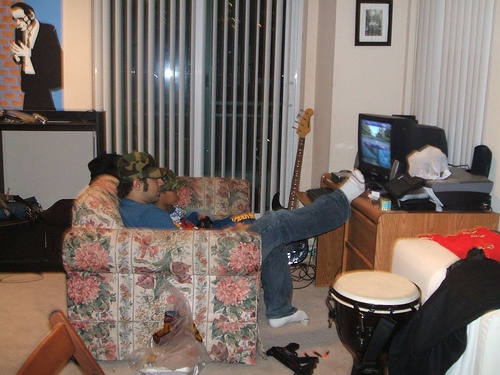}%
\hfill%
\includegraphics[width=0.33\linewidth]{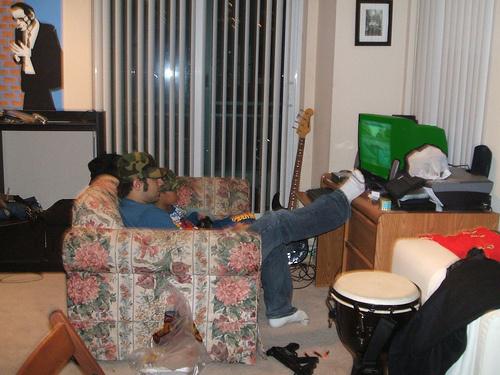}%
\hfill%
\includegraphics[width=0.33\linewidth]{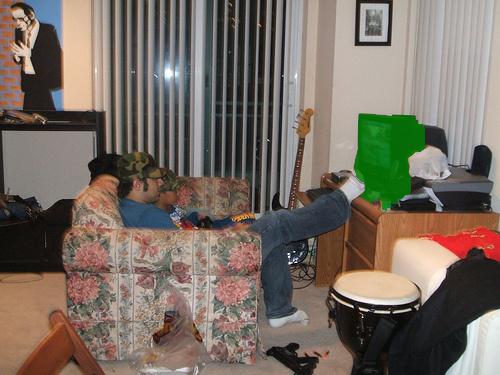}%
\end{subfigure}%
\\%
\begin{subfigure}{\figwww\linewidth}%
\includegraphics[width=0.33\linewidth]{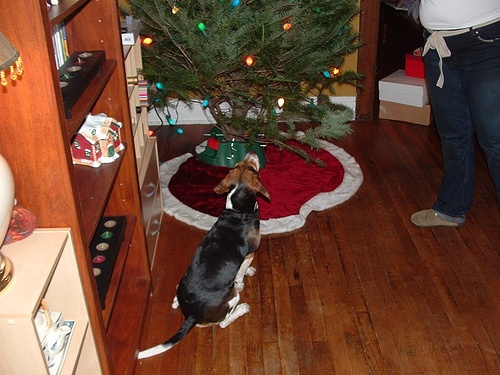}%
\hfill%
\includegraphics[width=0.33\linewidth]{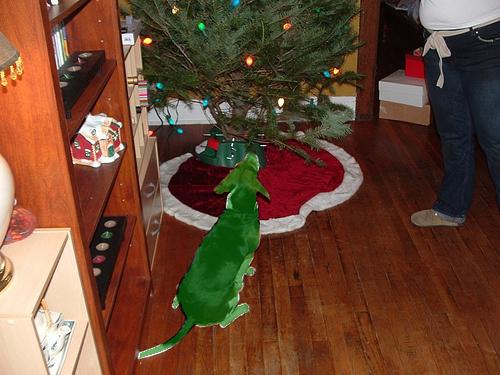}%
\hfill%
\includegraphics[width=0.33\linewidth]{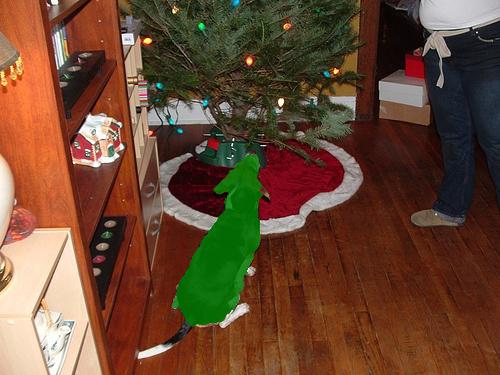}%
\end{subfigure}%
\\%
\begin{subfigure}{\figwww\linewidth}%
\includegraphics[width=0.33\linewidth]{figures/success/image4.jpg}%
\hfill%
\includegraphics[width=0.33\linewidth]{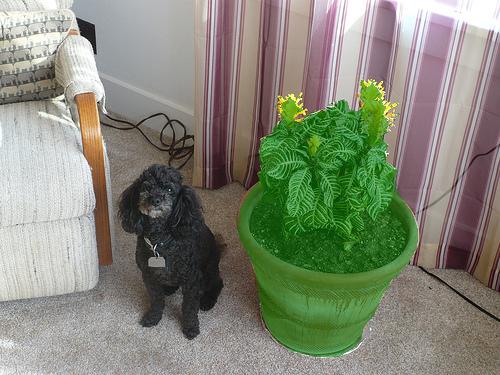}%
\hfill%
\includegraphics[width=0.33\linewidth]{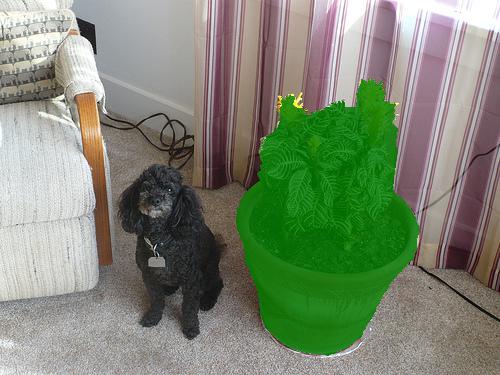}%
\end{subfigure}%
\caption{Left: original image; Middle: ground truth segmentation; Right: our resulting segmentation mask; Accurate segmentations for \textit{aeroplane}, \textit{tv-monitor}, \textit{dog} and \textit{potted-plant} (more results in supplementary material)}
\label{fig:cases1}
\vspace{-15pt}
\end{figure}

\begin{figure}[t]
\includegraphics[width=0.8\linewidth, trim={0.3cm 0 15.5cm 0},clip]{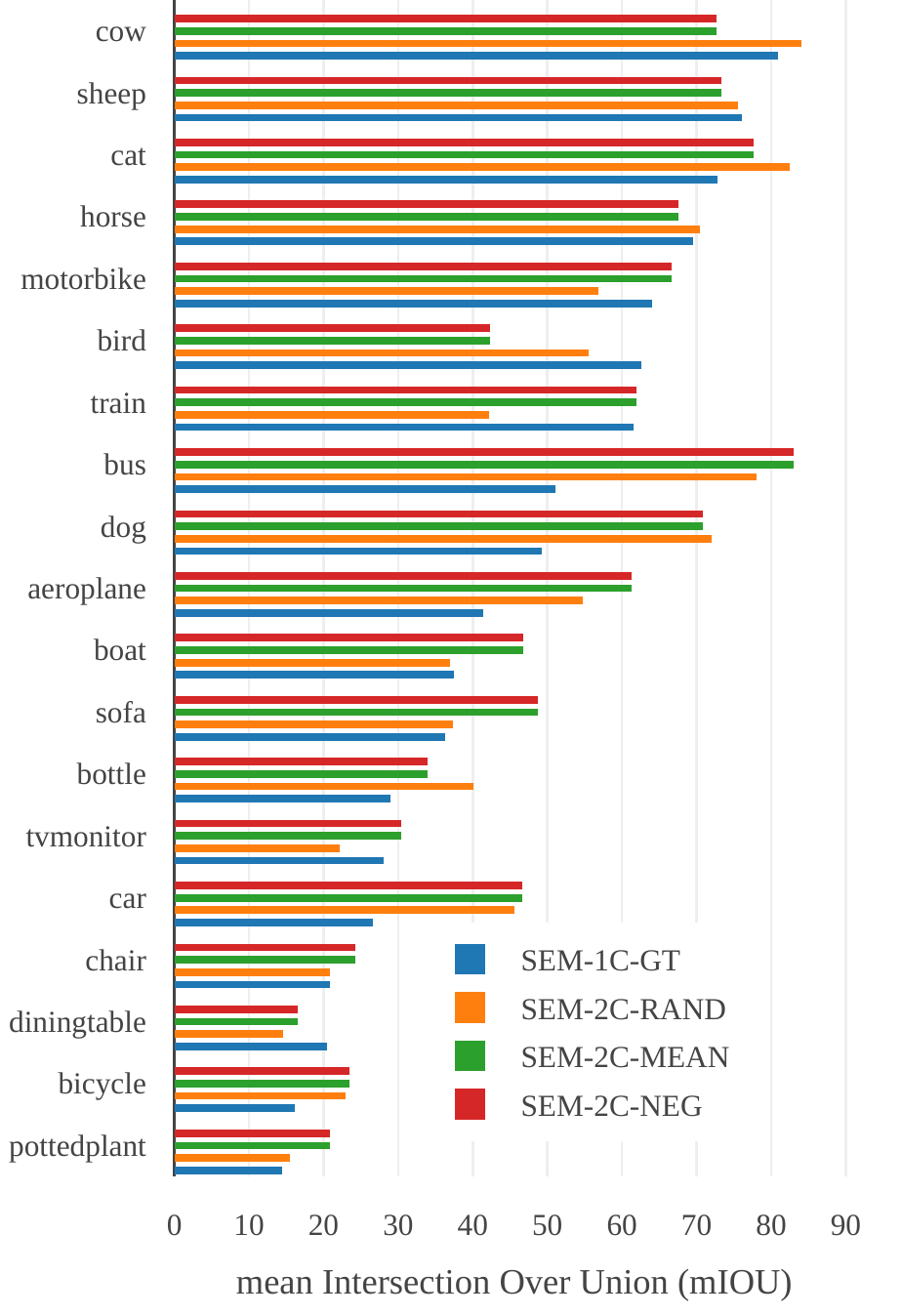}
\caption{Per-label mIOU of variants of our semantic segmentation models. \textit{1-channel} references to a model that uses one saliency map derived from the target label. \textit{2-channel-random}, \textit{2-channel-mean}, \textit{2-channel-proposed} refer to using a random label, the top 5 labels, and negative labels (\ref{model:seg:heat}) for the second saliency map}
\label{fig:perlabel}
\vspace{-20pt}
\end{figure}

The results of the comparison are depicted in Table~\ref{table:full_results_k1}, per-label results in Table~\ref{fig:perlabel}, and samples in Fig.~\ref{fig:cases1}. Even without any additional human annotations as support set, our proposed method (SEM-2-C-NEG) outperforms the work of \cite{zhang2018sg} on three out of four partitions, and achieves an overall mIOU of 50.3\%, a 4\% improvement over prior one-shot work ~\cite{zhang2018sg} that achieved an overall mIOU of 46.3\%. 

On three out of the four partitions introduced in~\cite{shaban2017one} we outperform previous state-of-the-art, without any human-provided images at test time. Our technique enables semantic segmentation models to extract salient visual cues from noisy saliency maps that segmentation networks are able to leverage to accurately segment out new labels at test time. Interestingly, we obtain better results from noisy self-generated saliency maps on the same image than~\cite{zhang2018sg} using human-generated ground truth labels on different images.

We observe that performance of the model varies significantly per-partition and per-label. We would like to better understand whether this is caused by the generated saliency maps or the segmentation model that uses them. Therefore, we train an \textit{oracle} segmentation model that uses the ground truth segmentation mask as positive saliency map. The oracle achieves an average mIOU of 95\% mIOU, indicating that a large part of the errors in our model would be alleviated by better activation maps. However, the oracle does not reach 100\% mIOU which points out that some errors are caused by the class-agnostic segmentation model. Interestingly, the oracle model and our proposed approach mostly suffers on the same target labels: While the median mIOU is over 97\%, the oracle model only achieves a 77\% mIOU on \textit{bicycle}, 88\% mIOU on \textit{pottedplant}, 94\% on \textit{chair}, the three worst performing classes of our model. This indicates that these classes are the most difficult to segment for the class-agnostic segmentation model.

\begin{figure}[t]
\centering
\begin{subfigure}{\figwww\linewidth}%
\includegraphics[width=0.33\linewidth]{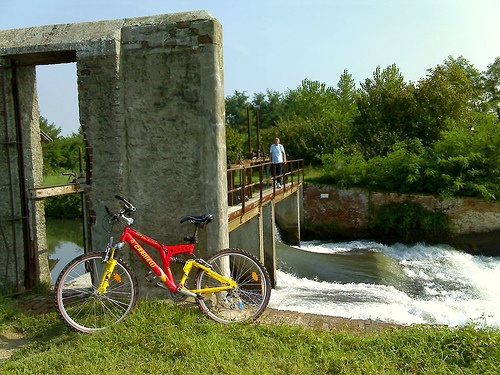}%
\hfill%
\includegraphics[width=0.33\linewidth]{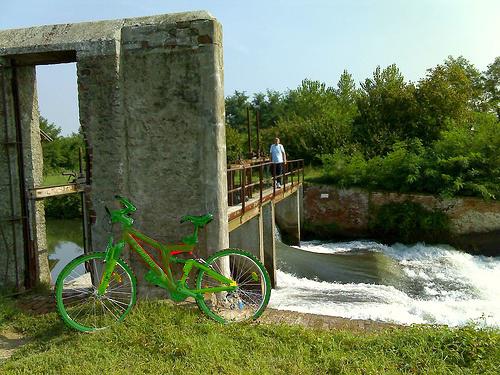}%
\hfill%
\includegraphics[width=0.33\linewidth]{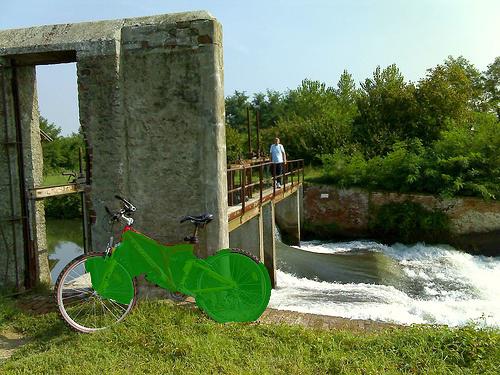}%
\end{subfigure}%
\\%
\begin{subfigure}{\figwww\linewidth}%
\includegraphics[width=0.33\linewidth]{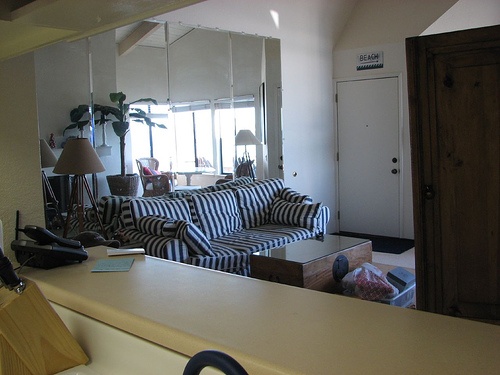}%
\hfill%
\includegraphics[width=0.33\linewidth]{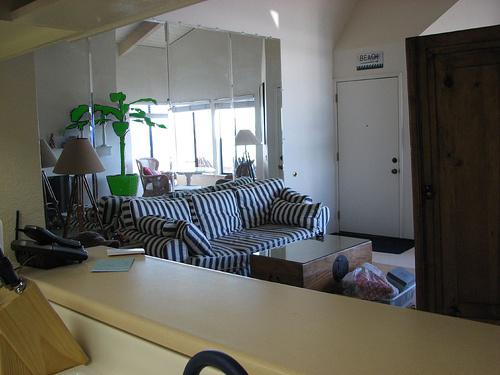}%
\hfill%
\includegraphics[width=0.33\linewidth]{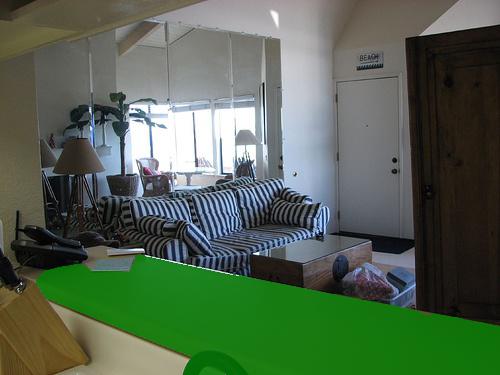}%
\end{subfigure}%
\\%
\begin{subfigure}{\figwww\linewidth}%
\includegraphics[width=0.33\linewidth]{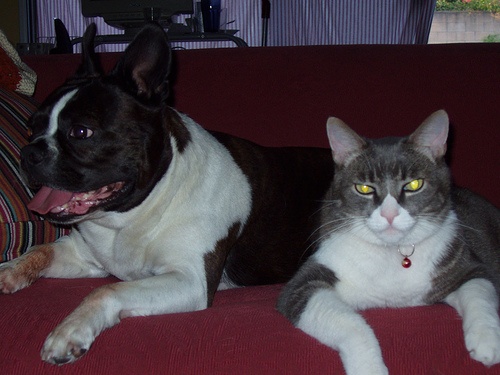}%
\hfill%
\includegraphics[width=0.33\linewidth]{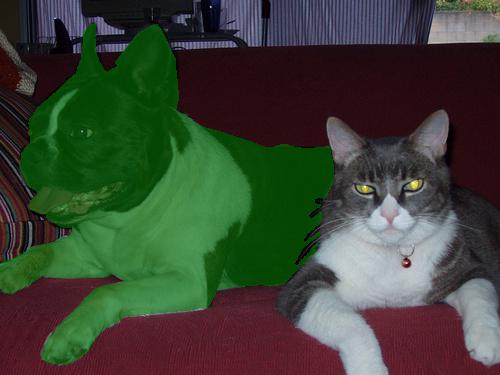}%
\hfill%
\includegraphics[width=0.33\linewidth]{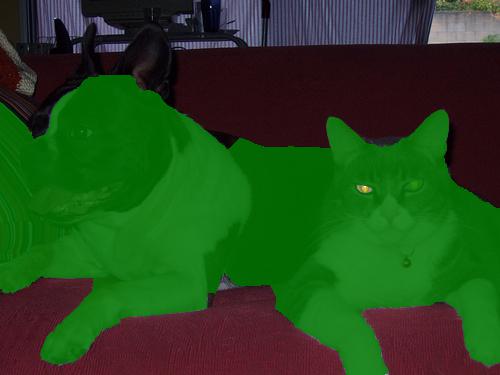}%
\end{subfigure}%
\caption{Left: original image; Middle: ground truth segmentation; Right: our resulting segmentation mask; Failure cases for \textit{potted-plant}, \textit{dog} and \textit{bicycle}. The top row shows incorrect object localization, the middle row shows semantically-similar object conflation, and the bottom row shows GrabCut postprocessing errors}
\label{fig:cases2}
\vspace{-20pt}
\end{figure}

We visually inspect failure cases from our model, observing the differences between the likelihood images and the final semantic masks, and find out that the majority of failure cases are due to either 1) GrabCut postprocessing errors (Fig.~\ref{fig:cases2}, first row), 2) object detection errors (Fig.~\ref{fig:cases2}, second row), or 3) attending on irrelevant but semantically related objects (Fig.~\ref{fig:cases2}, third row). We observe that the performance of GrabCut especially degrades when segmenting thin and/or occluded objects that are sparsely connected such as \textit{bicycle}, \textit{pottedplant}, and \textit{chair}, an avenue for improvement in future work. 




\begin{figure}[t]
\centering
\begin{subfigure}{\figwww\linewidth}%
\includegraphics[width=0.33\linewidth]{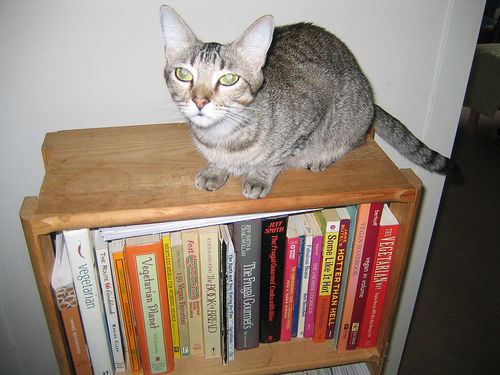}%
\hfill%
\includegraphics[width=0.33\linewidth]{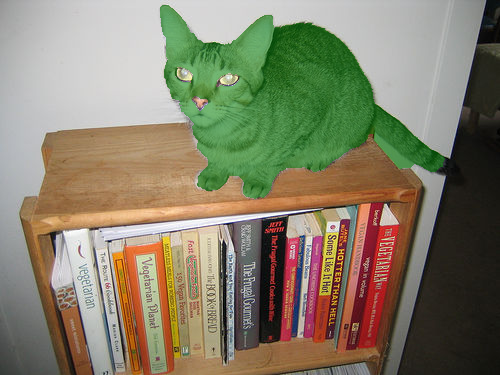}%
\hfill%
\includegraphics[width=0.33\linewidth]{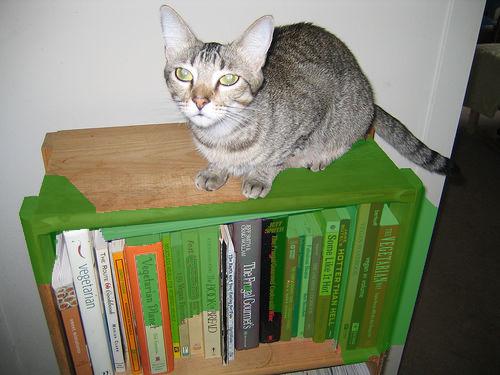}%
\end{subfigure}%
\\%
\begin{subfigure}{\figwww\linewidth}%
\includegraphics[width=0.33\linewidth]{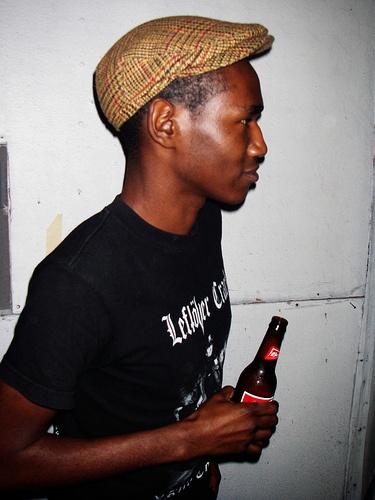}%
\hfill%
\includegraphics[width=0.33\linewidth]{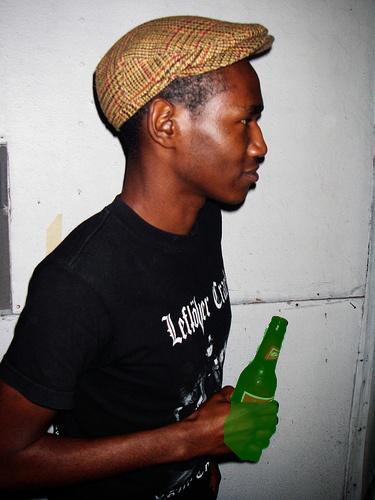}%
\hfill%
\includegraphics[width=0.33\linewidth]{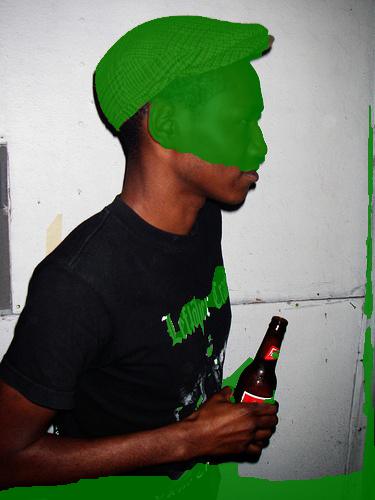}%
\end{subfigure}%
\caption{Qualitative results for our ablation studies. The top row shows results for a 1-channel model with saliency maps from target label \textit{cat} (middle) vs. random label \textit{banana} (right) vs. target label  (right); bottom row shows results for a 2-channel model (middle) vs 1-channel model (right) for target label \textit{bottle}.
}
\label{fig:ablationstudies}
\vspace{-15pt}
\end{figure}

\subsection{Role of Saliency Maps and Semantics}
In the third set of experiments, we analyze the importance of saliency maps and language semantics in our segmentation model. We use the same train setup as in \ref{ss:aolp}, and train three variants of a modified DeepLabv3 network. The first variant is a foreground/background classifier with no saliency maps as input (SEM-0C-NONE), the second uses a saliency map generated from 5 random labels (SEM-1C-RAND), and the last variant uses saliency map generated from a target label (SEM-1C-GT, section \ref{model:seg:heat}).
\label{ss:rsmas}
\paragraph{Improvement from saliency maps:}

In contrast to prior work ~\cite{rakelly2018conditional}, our proposed approach significantly outperforms a foreground-background baseline. A model with one saliency map from the target label outperforms the baseline on three out of four partitions, with an overall improvement of 5.6\% mIOU, while a model with two saliency maps outperforms the baseline across all partitions, and by 14\% mIOU overall. In contrast to a two-saliency-map model, a single-saliency-map model performs worse on partition 0 and only moderately better on partition 1 than the baseline. We hypothesize this is due to the noisy nature of salience maps; for semantically ambiguous labels such as \textit{car} or \textit{dog}, and small objects such as a \textit{bottle}, a single saliency map may pick up spurious objects in the image and thus misdirect the segmentation model towards an incorrect object.

\paragraph{Contribution from language semantics:}

We now analyze the contribution of the language-semantic mapping of the target labels to proxy labels in the pretrained model. We compare the results of that uses the target label (SEM-1C-GT) against a model that uses saliency maps from random labels (SEM-1C-RAND). We find that a model with saliency maps from random labels performs significantly worse across all partitions than a model with the target label. As illustrated in Fig.~\ref{fig:ablationstudies}, saliency maps from a random label often contain a greater degree of noise and focus on spurious pixels. This leads to less confident outputs from the segmentation network and incorrect segmentations after GrabCut postprocessing. However, as shown in the following section, negative saliency maps from random/distracting labels provide useful semantic information for the model, and performance improves significantly if combined with positive saliency maps from the target label.

\subsection{Two vs. One Saliency Map}
\label{ss:plac}

In the fourth set of experiments, we compare variants of our two-saliency-map model to a model that only uses a positive saliency map. 
In our two-saliency-map model, we experiment generating the negative saliency map with a random label, the mean of the top 5 labels predicted by the classifier for the target image, or the target negative labels (normal model). 

The per-partition results of our experiment are reported in Table~\ref{table:full_results_k1} and per-class results in Table~\ref{fig:perlabel}. The second saliency map provides complementary semantic information to the first map that our model successfully leverages. 

We observe an overall mIOU improvement of 8.3\% using two saliency maps over a one-saliency-map model. The largest performance gains are in the classes \textit{car}, \textit{bus}, \textit{dog}, \textit{aeroplane}, \textit{bottle}, and \textit{sofa}, that all have an average mIOU improvement of over 20\% as seen in Fig.~\ref{fig:perlabel} and exemplified in Fig.~\ref{fig:ablationstudies}, bottom. These objects often occur in cluttered environments with multiple objects in the scene. Adding a second saliency map improves both \textit{bus} and \textit{car} by a very large margin because 30.2\% of images with buses have cars in them. This shows that our method can disambiguate multiple objects in the same scene. 

Surprisingly, there is a small degradation for some classes when using two channels. The largest performance degradations are in the classes \textit{bird} and \textit{train}. Images that have these target labels depict typically only the target object in the image. This leads to positive and the negative saliency maps that are very similar to each other, leading to the degradation in performance when using both maps.

Our method is still unable to perform well in complex indoor scenes, as 4 of the 5 worst performing target labels are indoor objects. A large fraction of objects in indoor scenes are semantically related and co-occur with one another: 59.9\% of images with a \textit{dining table} have \textit{chairs}, and 21.8\% of images with a \textit{chair} have a \textit{sofa}. This poses a problem to dissambiguate them from the noisy saliency maps. The segmentation of multiple indoor objects is thus a promising avenue for improvement in future work.

